# Improving Condition- and Environment-Invariant Place Recognition with Semantic Place Categorization

Sourav Garg[1], Adam Jacobson[1], Swagat Kumar[2] and Michael Milford[1]

*Abstract*— The place recognition problem comprises two distinct subproblems; recognizing a specific location in the world ("specific" or "ordinary" place recognition) and recognizing the type of place (place categorization). Both are important competencies for mobile robots and have each received significant attention in the robotics and computer vision community, but usually as separate areas of investigation. In this paper, we leverage the powerful complementary nature of place recognition and place categorization processes to create a new hybrid place recognition system that *uses place context to inform place recognition*. We show that semantic place categorization creates an informative natural segmenting of physical space that in turn enables significantly better place recognition performance in comparison to existing techniques. In particular, this new semantically informed approach adds robustness to significant *local* changes within the environment, such as transitioning between indoor and outdoor environments or between dark and light rooms in a house, complementing the capabilities of current condition-invariant techniques that are robust to *globally consistent change* (such as day to night cycles). We perform experiments using 4 benchmark and new datasets and show that semantically-informed place recognition outperforms the previous state-of-the-art systems. Like it does for object recognition [1], we believe that semantics can play a key role in boosting conventional place recognition and navigation performance for robotic systems.

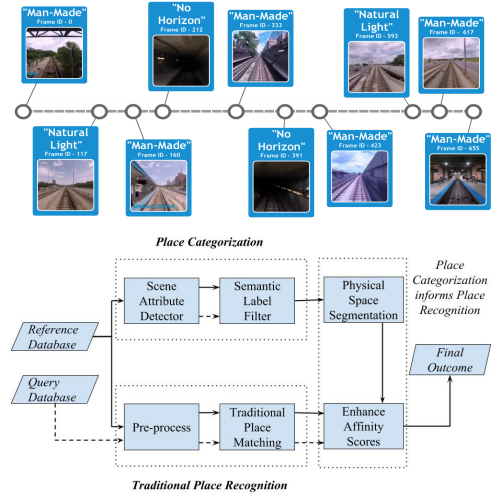

Fig. 1. The top row shows semantic labels corresponding to different segments of the environment in CTA-Rail dataset. The bottom figure shows a block diagram showing the flow of semantic information from the place categorization module to the place recognition module for improving place matching scores.

## I. INTRODUCTION

The problem of traditional place recognition typically focuses on recognizing specific locations in the world stored within a database of "places". This form of place recognition is very powerful, enabling localization on very large scales [2] and during difficult day and night traverses of an environment [3]. The problem of place categorization is similar to the place recognition problem, where environments are evaluated to determine the type of place from a database of place types.

The place recognition problem has historically been solved for ideal environmental conditions [4], [2] until recently with a significant interest towards an all weather [5], all season [6], [7], all times-of-day [8], that is, all-condition place recognition [3], aiming for long-term localization and mapping. The research has further led to a creation of massive 1000 km dataset [9] having repeated traversals of different places over a time period of one year, therefore exhibiting variety of conditional and structural changes in the environment across the traversals.

The variations in environmental conditions for visual place recognition have so far primarily been explored for global changes in conditions across different traversals of the same route. For example, a single route traversal at night exhibits only night conditions throughout the traverse; similarly, a route traversed during sunny weather doesn't exhibit other weather conditions in the same traverse. In many real world situations, lots of local changes in environmental conditions are encountered, especially by long-term, seamless navigation systems. Such local changes can either be moderate like moving from an indoor environment to an outdoor one, or can be extreme like transiting from an artificially-lit underground car-parking to an open outdoor road-network at night; or entering a house during daytime and navigating in the dark before turning on the lights.

In this paper, we study the effects of local condition variations in the environment on a state-of-the-art place recognition system which is globally condition-invariant. We also present a method to deal with such condition variations within or across the traversals and develop a novel framework to incorporate semantic labels and place categorization results to inform and improve place recognition place estimates by dividing the environment into meaningful segments (as seen in Fig.1). Once a place is categorized, we leverage the SeqSLAM [3] framework to perform place recognition,

[1]The authors are with Australian Centre for Robotic Vision, Queensland University of Technology, 2 George St, Brisbane, Australia
[2]The author is with Innovation Labs, Tata Consultancy Services, New Delhi, India

implementing a novel dynamic weighting scheme, biasing place matches with similar place characteristics and place categorization results.

We evaluate our proposed approach on four real world datasets. These datasets are sufficiently diverse to cover variations in route length, viewpoint, type of motion, and global and local environmental conditions. We show that the performance of proposed approach improves upon whatever performance the baseline approach has, by incorporating place categorization information.

The paper proceeds as follows: In Section 2, we review literature with a focus on place recognition and place categorization, Section 3 presents our approach describing the CNN place categorization framework, outlines our place recognition framework and the proposed technique for combining the two frameworks to produce superior place recognition results. We present the experimental setup in Section 4, and results of multiple levels of evaluation, along with a discussion in Section 5. Section 6 concludes the paper and highlights areas of future work.

## II. RELATED WORK

In this section, we review current research in the areas of place categorization and place recognition. We specifically focus on place recognition, semantic mapping and place categorization frameworks.

### A. Place Recognition

Visual place recognition leverages a visual map of the environment and compares visual information, typically from a camera sensor, with the map data to determine the current location of the camera within the map. There are many techniques which have been proposed to solve this problem of determining where an image has been taken within an environment. Typically, these approaches leverage single frame matching to determine the location of the camera in the environment. The key goal of place recognition frameworks is to separate places in the environment and highlight the unique attributes or features which uniquely describe individual locations in the environment [2].

There have been many attempts to improve performance of place recognition systems. This has included the inclusion of temporal information, fusing multiple sensory modalities and implementing unique preprocessing steps to improve localization capabilities like shadow removal techniques [10].

Temporal information is generally incorporated into the place recognition framework [3], integrating place hypotheses over small distances to accrue evidence and improve place recognition performance.

Furthermore, the introduction of unique sensor preprocessing techniques to improve sensor data for place recognition has also been explored. Frameworks utilizing techniques for shadow removal [10] or the introduction of illumination invariant color spaces [11] to remove temporal or environmental changes from images to improve localization.

A significant amount of work has also been done to make place recognition algorithms robust to global changes in environmental conditions [12], [6], but none of those mentions about robustness towards local changes in environmental conditions.

### B. Place Categorization

Place categorization systems are an extension of the place recognition problem and attempt to attach semantic meaning to particular places in an environment; attempting to utilize labels from a training set like indoor, outdoors, kitchen, office and bedroom to categorize the location within which an image was taken. These frameworks are powerful as they facilitate generalization of room labels to different environments, for example identifying a bedroom within an unexplored house, potentially enabling robotic platforms to perform generic tasks in unknown environments by recognizing the type of place [13].

There have been a number of works which attempt to imbue traditional SLAM architectures with the ability to semantically label locations in an environment [1]. These types of frameworks utilize the place categorization labels to provide information about a space, for example, identifying a location to be a kitchen, but these labels are not utilized in the process of generating the map or performing place recognition or localization.

In a recent work [14], authors develop a method to generate different categories of environments from a large available reference database for place recognition in order to reduce the search space for matching places. They basically segment the overall physical space into categories of similar environments within the place recognition system and do not use any semantic place categorization.

Place categorization systems have been leveraged in previous work to improve object detection and classification, enabling reduction of the object search space and improvement in object recognition performance [15].

However, there has been no prior work utilizing place categorization information to improve place recognition place estimates under the influence of local and global changes in environmental conditions.

## III. PROPOSED METHOD

The proposed approach has three main components: place categorization, physical space segmentation and place recognition as depicted in Fig. 1, with semantic information flowing from the former to the latter to generate the final place match estimate. Our core contribution is the development of a technique to utilize place categorization information to improve place recognition performance. In order to achieve this, we use the semantic labels to divide the physical space into different regions based on its appearance, that is, semantic scene attributes. These segmented regions are then used in the place recognition module for biasing place matches that lie in a particular semantic region. We use CNN model VGG16-places365 [16] pre-trained on the Places365 database [17] for labeling reference database frames with most probable scene attributes [18]. We use SeqSLAM [3] for showing improved place recognition using semantic information by appropriately enhancing the image matching scores.

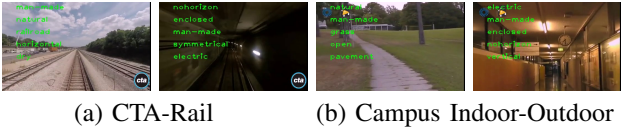

(a) CTA-Rail      (b) Campus Indoor-Outdoor

Fig. 2. Images with top-5 most probable semantic labels out of 102 scene attributes from the reference databases of two of the datasets.

### A. Place Categorization

The pre-trained CNN model classifies an image with probabilities associated with each of the 365 place categories. It is also made to predict the most probable scene attributes (out of 102 attributes trained on the SUN database [18]) using one of its fully-connected layers ('fc7'). We use these scene attributes and their associated probabilities to post-process the image labels for semantic segmentation of datasets. The predicted scene attributes for some of the images from datasets used in this paper are shown in Fig. 2. The classification is performed only on the reference database. The semantic labels as obtained are used to temporally divide the reference image sequence into different segments.

### B. Physical Space Segmentation

The place categorization module provides semantic labels for each reference image ranked according to the probabilities associated with those labels. In order to segment the reference database, a unique label corresponding to each image is required while taking into consideration the temporal nature of the input and avoiding any transient errors produced by place categorization module. This is achieved by using a Hidden Markov Model (HMM) [19], [20], where we estimate the model parameters and the hidden state corresponding to each image in the reference image sequence using the semantic label probabilities as the observed variable for that image.

The sequence of semantic labels feature vector corresponding to the reference database having $T$ number of images is represented as a random variable $X = (x_1, x_2 \ldots x_T)$ and the hidden variables are denoted as a random variable $Z = (z_1, z_2 \ldots z_T)$, where $z_t$ at a given time instant $t$ can belong to one of the $N$ hidden states. It is assumed that given the $z_{t-1}$, $z_t$ is independent of previous hidden variables and current observation $x_t$ only depends on current hidden state $z_t$. Hence, the state transition probability matrix, represented as $A$ and initial state distribution $\pi_i$ is given as:

$$A = \{a_{ij}\} = p(z_t = j | z_{t-1} = i) \quad \forall i, j \in [1, N] \quad (1)$$

$$\pi_i = p(z_1 = i) \quad (2)$$

The probability of an observation at time $t$ for being in state $i$ is defined as:

$$b_i(x(t)) = p(x(t)|z_t = i) \quad (3)$$

Our objective is to find the hidden state sequence of the model, that is, the desired unique labels for the reference image sequence. This is given by the posterior probability of the state sequence:

$$p(Z|X, \theta) = \frac{p(X, Z|\theta)}{P(X|\theta)} \quad (4)$$

where $\theta = (A, b_i(x(t)), \pi)$ are the parameters of the model and

$$p(X, Z|\theta) = \pi_{z_1} \prod_{t=1}^{T-1} a_{z_t z_{t+1}} \prod_{t=1}^{T} b_{z_t}(x(t)) \quad (5)$$

$$P(X|\theta) = \sum_Z p(X, Z|\theta) \quad (6)$$

The final labels for the reference images, represented as $L_t$, are obtained after estimating the parameters $\theta$ of the model:

$$L_t = \arg\max_i Z_t(i) \quad \forall i \in [1, N] \quad (7)$$

The input feature vector, that is, the observation $x(t)$, is the output response vector of the place categorization module with size 1x102, where 102 dimensions represent the probability associated with each of the scene attributes. The feature vector is normalized to the range $[0, 1]$ before feeding into the HMM. The parameters $\theta$ of the model are determined using Baum-Welch algorithm [21] and the most likely hidden state sequence is obtained. The implementation of HMM used for this work is available here [20]. The number of hidden states, that is, $N$ is empirically determined for the datasets used in the paper, though there are ways to determine $N$ using cross-validation [22] or by using Infinite HMM [23], and is not the focus of our work. Fig. 1(a) shows the images and their semantic labels at the segmentation transition points for one of the datasets used in this paper.

### C. Place Recognition

In general, a place recognition system comprises of a pre-processing stage, then a method to calculate affinity scores between database places and the query, and finally a decision module for generating the best matching pairs, as seen in Fig. 1.

*1) Sequence-based place matching:* In addition to the above mentioned place recognition pipeline, sequence-based recognition methods exploit the temporal information inherent in this problem. SeqSLAM [3] is known to work well in challenging environmental conditions and is able to recognize places despite seasonal, weather or time of day variations. The recent advanced methods [24], [25]etc. inspired from SeqSLAM further improve the state-of-the-art for place recognition. In this paper, we use the vanilla approach to show the performance improvement of a place recognition system, under the influence of variations in the surrounding environment, with the help of semantic information associated with those places. The detailed methodology of SeqSLAM can be referred to in [3].

SeqSLAM performs place recognition using Sum of Absolute Difference (SAD) scores represented as $D$ between preprocessed reference and query images. The preprocessing step involves down-sampling of image to size $S_x$ and $S_y$ and patch normalizing it with a fixed square window of side length $P$.

$$D_i = \frac{1}{S_x S_y} \sum_{x=0}^{S_x} \sum_{y=0}^{S_y} |p_{x,y}^j - p_{x,y}^i| \quad (8)$$

where $p_{x,y}^i$ and $p_{x,y}^j$ are the pixel intensities of patch normalized reference and query images.

The difference vector obtained for each query image undergoes neighborhood normalization within a sliding window of size $R$, also termed as neighborhood normalization zone width. The neighborhood normalized difference for a given query image, $\hat{D}_i^R$ is calculated using the local mean difference $\bar{D}_i^R$ and local standard deviation $\sigma_i^R$.

$$\hat{D}_i^R = \frac{D_i - \bar{D}_i^R}{\sigma_i^R} \quad (9)$$

The neighborhood normalized SAD matrix is then searched for local image sequence trajectories of length $d_s$, within a limited range of velocities, originating from each of the reference image. The sequence trajectory with the best score is then selected using a trajectory uniqueness threshold $\mu$.

*2) Localized and semantically-informed matching:* The neighborhood normalization of place matching scores within the window $R$, as calculated in Eq. 8 and 9, reflects the emphasis on matching a local physical region of the environment, instead of finding a global minima. The parameter $R$ represents the span of environment, where the matching scores can be locally enhanced. Our aim is to pre-define these spatio-temporal regions of the environment that share similar semantic labels.

The segmentation of the dataset as described in earlier sections using HMM separates the physical space into regions with similar environmental conditions. As shown in Fig. 1, in general, a place recognition system can use the semantic information from the place categorization module to enhance its affinity scores for matching places. Instead of arbitrarily choosing the neighborhood for the reference image as in Vanilla SeqSLAM method, we propose to use the neighborhood regions obtained using labels $L_t$ generated by HMM. The segmented regions are denoted as a set of pairs $R'_t$:

$$R'_t = \{(i,j) \mid t \in [i,j] \text{ and } L_k = L_{k+1} \quad \forall k \in [i,j] \\ \forall i,j \in [1,T]\} \quad (10)$$

where $t$ iterates over all the reference images and a pair in $R'_t$ defines the lower and upper limit of the segmented region within the reference database. The neighborhood normalization equation (9) is therefore updated as below:

$$\hat{D}_i^{R'_i} = \frac{D_i - \bar{D}_i^{R'_i}}{\sigma_i^{R'_i}} \quad (11)$$

Fig. 5 shows the method described in this section for choosing $R$. The incorporation of segmented regions based neighborhood normalization as shown above, makes sure that different environmental conditions encountered within a traversal are handled separately for finding a local best match.

## IV. EXPERIMENTAL SETUP

The experiments are performed using four different datasets described in following subsections. The image classification for place categorization is performed off-line as a preprocessing step and all the experiments are conducted using Dell Latitude E7450 Intel Core i7-5600 CPU @ 2.6 GHz x 4 processor having 16 GB RAM and running Ubuntu 14.04.

### A. Datasets

The four datasets used in the experiments are practical application scenarios for visual SLAM and seamless navigation systems and exhibit diversity with respect to the (a) route length - varying from small campus traversals to large rail routes; (b) condition variations within the traversal - medium variations like transiting from open-space to enclosed ones, and extreme variations by making such transitions from bright to dark space or vice-versa; (c) condition variations across the traversals - from none to day and night; (d) amount of motion - varying from pedestrian motion, to motorbike and to trains; (e) viewpoint - from slight variations due to gait and jerks on bike, to deliberate lateral offsets along the footpath. The aerial view of the datasets along with some sample images is shown in Fig. 3

*1) CTA-Rail:* The CTA-Rail (Chicago Transit Authority) dataset comprises two videos traversing a 23 km railway route (Blue Line, Forest Park to O'Hare), recorded once in 2014 and then in 2015 [26], available online. A single camera is placed at the head of the train facing forward towards the railway track. The videos comprise scenes from train stations platforms, subway station platforms, subway tunnels, and railroad tracks within highways and urban areas. The raw videos are approximately 73 and 84 minutes in duration with 132670 and 149090 frames respectively. We used the 480p version of the video and processed every 200th frame for all the experiments. The resultant reference and query databases have therefore 656 and 738 image frames respectively.

*2) Campus Indoor-Outdoor:* The Campus Indoor-Outdoor dataset comprises of two videos with repeated traversal of a part of Ulm University of Applied Sciences' campus from an outside lawn to an inside corridor [27]. The videos have been recorded using a hand-held device with single camera and exhibit jerky motion with huge motion blur. The raw videos are cropped to remove the comments at the bottom and an overlaid navigation display on the right side. The videos are also snipped from beginning so that the starting point is aligned in both the datasets. The datasets comprise of scenes from outside the campus, with trees, grass and pavement, and from inside the campus, traversing through entrance hall, staircase, lobby and corridor. The reference and query database is processed by using every 10th image and therefore uses 355 and 300 frames respectively.

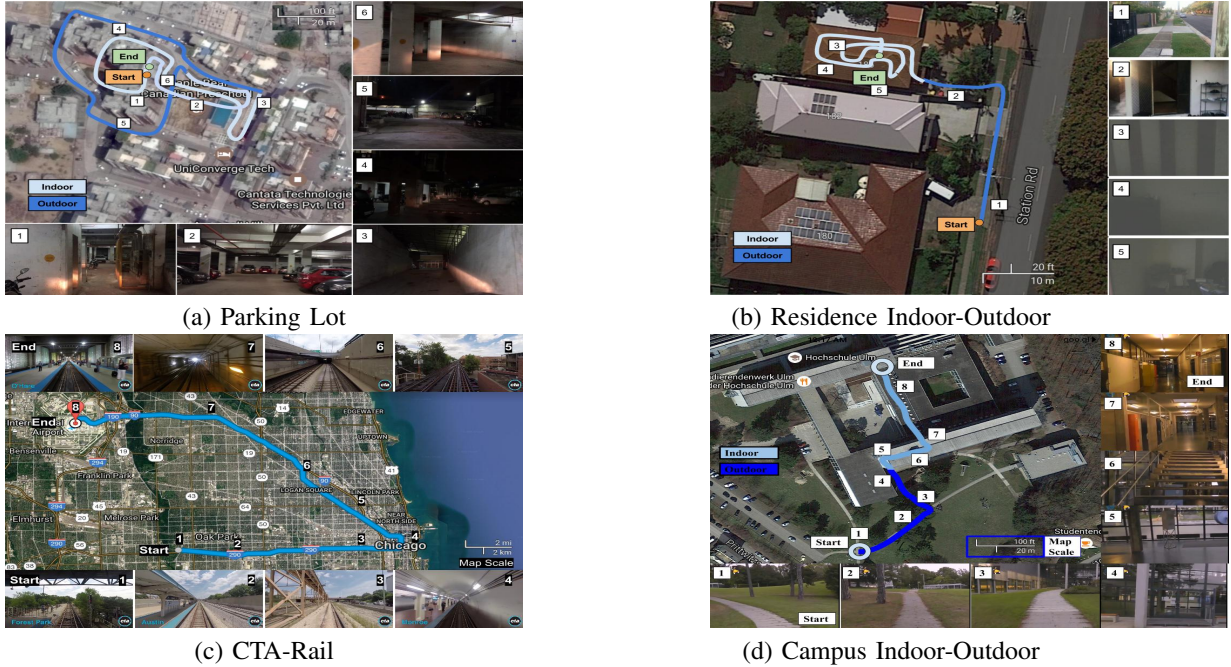

(a) Parking Lot  (b) Residence Indoor-Outdoor
(c) CTA-Rail  (d) Campus Indoor-Outdoor

Fig. 3. Aerial View of all the four datasets with sample images. The images captured at different time instants throughout the traverse show how the transitions within the environment take place. (Source - Google Maps)

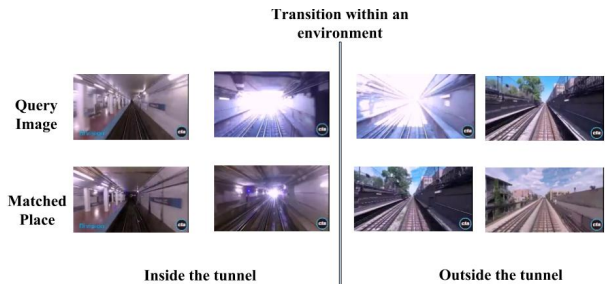

Fig. 4. Matched places at the point of transition within an environment for CTA-Rail dataset where environment changes from being inside the tunnel to outside the tunnel.

*3) Parking Lot:* The Parking Lot dataset is captured inside a society of residential buildings, traversing through the underground as well as open parking area. It comprises of two videos traversing the same path, once during daytime and then at night. The reference database exhibits transition from underground parking area having artificial lighting to open naturally lit space during day time, and the query database transits from the underground parking into the dark sky during night time with some street lights. This dataset hence possess variations in environmental conditions within as well as across the route traversals. The videos have been captured using hand-held mobile device while driving on a motor-bike and contain 6407 and 6396 image frames respectively for day and night traversal. We process every 20th frame, therefore processing 320 frames for each of the reference and query database.

*4) Residence Indoor-Outdoor:* The Residence Indoor-Outdoor dataset comprises of two traversals from outside of a house and then entering inside the house via corridors to the common area and then to the bedroom via stairs. The reference database was captured during daytime with good natural lighting outside the house and with minimum lighting inside the house. The query database was captured at night with street lights lighting the way outside of the house and adequate artificial lighting inside the house. Therefore, in this dataset as well, there are variations in the environmental conditions within and across the traversals. Moreover, the path traversed outside the house also exhibits a change in viewpoint in the two traversals, due to a deliberate lateral offset of around 1m while walking down the pavement. The videos are captured using hand-held camera. The reference and query database comprise of 2200 and 2180 image frames respectively and are processed by skipping 10 frames, therefore comparing 220 and 218 image frames.

*B. Ground Truth*

The place recognition ground truth for all the datasets was generated manually for intermittent frames and then interpolated for the rest of the image sequence. A query image is considered to be a true positive match for the reference image if its index lies within a range of 5 image frames from the ground truth index.

*C. SeqSLAM parameters*

The parameters for SeqSLAM used for all the experiments are shown in Table I.

## V. RESULTS AND DISCUSSION

We used maximum F1 score to measure changes in place recognition performance using the proposed approach. The trajectory uniqueness parameter (described in [3]), that is, the threshold for deciding a correctly matched place was varied to calculate precision-recall curve and maximum F1 score. The comparative results were generated between the proposed method and vanilla SeqSLAM for four real world datasets. In order to gain an in-depth understanding of the place recognition performance changes due to proposed approach, we used two parameters of SeqSLAM method, patch normalization window size ($P$) and neighborhood normalization zone width ($R$), to obtain performance trends. The results are as shown in Fig. 6, and the significance and effect of these parameters is discussed in subsequent section.

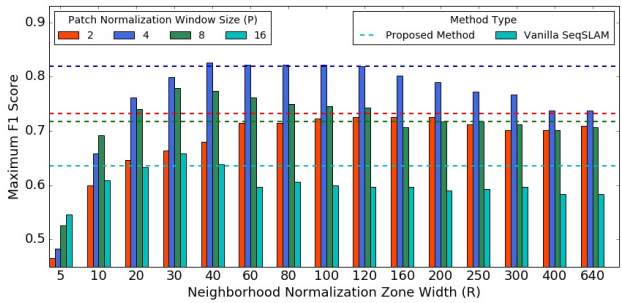

(a) Parking Lot

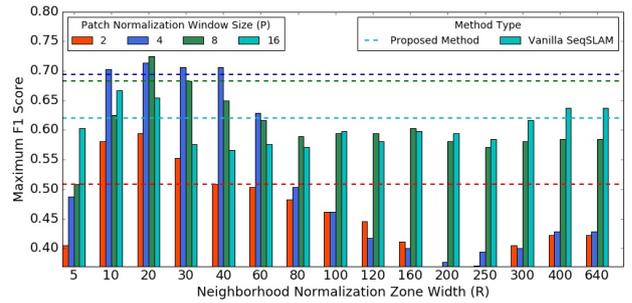

(b) Residence Indoor-Outdoor

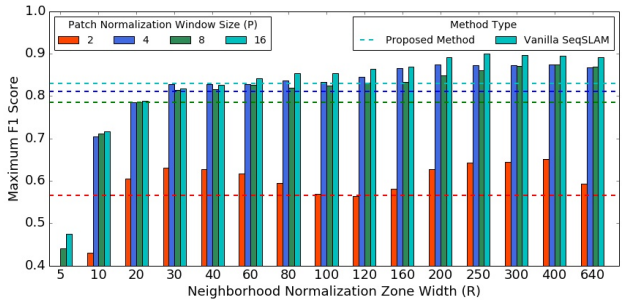

(c) CTA-Rail

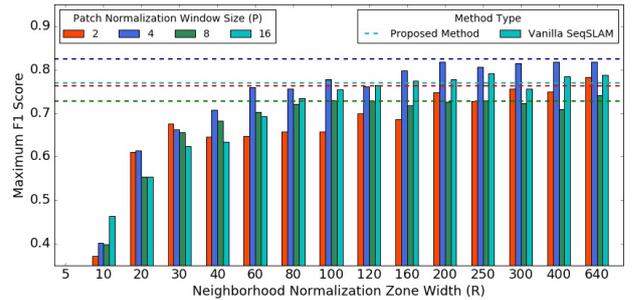

(d) Campus Indoor-Outdoor

Fig. 6. Performance charts showing maximum F1 Score w.r.t. different R (Neighborhood Normalization Zone Width) values. R is varied to the maximum value, that is, the size of reference database, after which maximum F1 score becomes constant. The patch normalization window size ($P$) parameter with value 4 happens to perform better as compared to others most of the times.

TABLE I
SEQSLAM PARAMETERS.

| | | |
|---|---|---|
| $S_x \mathbf{x} S_y$ | Image Down-sampling Size | 64x32 |
| $P$ | Patch Normalization Window Size | 2,4,8,16 |
| $O$ | Image Matching Offset Range | $\pm 10$ |
| $d_s$ | Sequence Length | 15 |
| $R$ | Neighborhood Normalization Zone Width | Varies from 5 to 640 |
| $V$ | Sequence Search Velocity Range | $(1 \pm 0.2)d_s$ |
| $\mu$ | Trajectory Uniqueness Threshold | Varied |

### A. Neighborhood Normalization Zone Width ($R$)

The neighborhood normalization zone width $R$ defines a temporal region around the reference image in order to find a local best match for the query image. As shown in Fig. 6, the proposed approach outperforms the vanilla method by adequately segmenting the reference image database and selecting the right temporal region for enhancing the place matching scores. The regions $R'$ being determined using semantics are independent of the parameter $R$, hence the performance measure is always constant w.r.t. it.

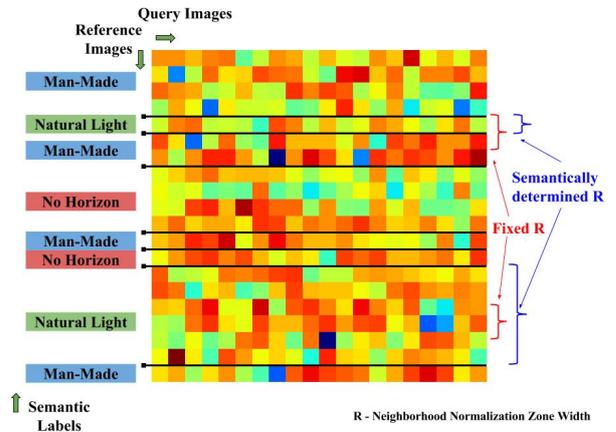

Fig. 5. The semantic segmentation of environment decides the normalization regions for better place recognition. The matrix represents the Sum of Absolute Difference score between reference and query images of CTA-Rail dataset. The black horizontal lines mark the transitions from one type of environment to the other. The red markers on the right show the fixed neighborhood normalization zone width (R) for SeqSLAM and blue markers refer to the proposed method for determining R.

Our proposed method performs consistently better than the vanilla approach for smaller values of $R$ for all the datasets. This is an expected outcome as a smaller temporal window around the reference image means spanning across very similar images, wherein normalization of matching scores creates local maximas and minimas within that region. As this process is repeated for all the reference images, the local extremas being very similar to each other cause inter-region redundancy and therefore add confusion, leading to false matches. Ideally, a temporal region around the reference image should be such that it spans across non-overlapping dissimilar images in the environment in order to correctly recognize a local match. This is achieved by our proposed method as it uses the semantically-segmented environment for creating adequate local temporal regions to effectively highlight the true matches.

It can be noted in the Fig. 6, that the performance of vanilla SeqSLAM for CTA-Rail and Campus Indoor-Outdoor dataset gets better with increasing the parameter $R$, but for the Parking Lot and Residence Indoor-Outdoor datasets, it achieves its peak performance and then starts to fall before becoming constant. This happens because of the fact that the former datasets exhibit moderate variations in conditions within the traversal and none across the traversals, whereas the latter exhibit extreme changes in condition both within and across the traversals.

A large normalization zone essentially means finding a global match in the entire reference database which gives rise to false matches as variations in environmental conditions across traversals become extreme. For example, in Residence Indoor-Outdoor dataset's first traversal (as shown in Fig. 3(b)), images in the beginning have been captured in broad daylight outdoor setting, which then transits to indoor environment with images captured in darkness of enclosed hall and bedroom. On the other hand, the images from second traversal initially pass through the outdoor environment at night under street light and then transit into indoor areas of the residence brightened by lamps and bulbs. This is also shown in Fig. 7, where 7, (a) shows the ground truth images, (b) shows mostly false matches that occur using vanilla SeqSLAM with $R$ value set to its maximum ($R = 640$), and (c) shows the correct matches using the proposed method. The false matches of vanilla SeqSLAM show that it finds a global minima that matches dark with dark and bright with bright rather than realizing that the condition-variant true match of query image lies in the other half of the reference database. This pitfall is generally avoided in SeqSLAM by using the local best match in an fixed-size window $R$, which is arbitrarily chosen and hence needs to be determined. We overcome the same in this paper by semantically segmenting the environment and determining the right value of the parameter $R$ corresponding to each reference image in the database in order to correctly recognize places despite extreme changes in environmental conditions *within* or *across* the traversals.

*B. Patch Normalization Window Size (P)*

The images used for finding SAD score are preprocessed by down-sampling them to the size of 64x32 and then patch normalized in order to counter the variations in environmen-tal conditions as proposed in SeqSLAM. Depending on the type of environment and corresponding imagery, the choice of patch normalization window size can lead to changes in performance. It is evident from Fig. 6, that the performance trends obtained by varying $R$ are similar for different values of $P$, but there are no specific trends for performance versus $P$ itself. In the experiments performed for current work, we found that for our proposed method, patch normalization window size of 4 for the given down-sampling image size performs better in most of the cases.

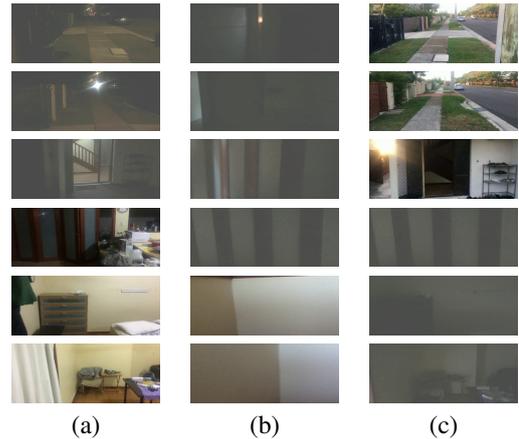

Fig. 7. (a) Ground truth images from the Residence Indoor-Outdoor dataset in a temporal order showing transition from nightly outdoor environment into bright indoor areas. (b) Matched places using Vanilla SeqSLAM mostly showing false matching of images having similar lighting conditions. (c) Matched places using proposed method showing correct place recognition. It also shows the transition in environment from broad daylight to dark indoor areas. Note: The images captured at night or in dark are shown here after manually brightening them for sake of visualization only.

*C. Viewpoint Variation*

The Residence Indoor-Outdoor dataset was captured with lateral offset of approximately $1m$ between the camera position in its reference and query database. Although the baseline performance is lower, due to SeqSLAM's limited ability to deal with viewpoint variation, the semantic segmentation improves the place recognition performance. This was done in order to show that the performance of proposed approach improves upon the baseline approach with whatsoever capabilities the baseline approach has. The research problem pertaining to developing condition as well as viewpoint invariant place recognition system has been looked into widely [24], [7] and is not the aim of our current work.

## VI. CONCLUSION AND FUTURE WORK

In this paper, we proposed a method to deal with significant changes in local and/or global conditions for a place recognition system by using semantic labels information to segment the environment into meaningful chunks. We showed how local condition variations clubbed with global condition variations can significantly hamper the performance of a frame-based condition-invariant place recognition system. We presented four real-world datasets depicting the practical situations where such local and global changes in environmental conditions occur to show that semantically-informed system outperforms previous state-of-the-art system.

The current work can be extended to a sophisticated visual SLAM system, where visual odometry and mapping can benefit too from the semantic information. The use of a local temporal region around a reference image is also encouraged from the fact that a large-scale place recognition or SLAM system will probably perform an initial coarse localization using semantic labels and then fine tune it for exact localization. However, in this work, we don't use any such filtering of places because state-of-the-art place categorization systems have not been trained on such a wide variety of datasets that would cover all the different environmental conditions for all the existing places in its training data. To enable the benefits of semantic place categorization to be universally applied or applied in a specific domain, place categorization deep nets will need to be trained with domain-relevant data. Here we have used semantic place information to inform one aspect of specific place recognition using place category segments to control the normalization zones for a method like SeqSLAM. However, there are other potential benefits to using semantics for example semantics about the nature of the place could also be used to determine what type of feature detector should be used, or what level of viewpoint invariance is required (for example in a self-driving car domain). Finally, it may be possible to close the loop back by using the specific place recognition information to fine tune (during training) or correct (during operation) the place categorization system, or ultimately to jointly recognize both the specific place and the category of place.